\documentclass[%
aip-cp,
 jmp,%
 amsmath,amssymb,
preprint,%
]{revtex4-1}

\usepackage{graphicx}

\usepackage{dcolumn}%
\usepackage{bm}%
\usepackage{neuralnetwork}
\usepackage{tikz}

\def\ltsima{$\; \buildrel < \over \sim \;$}
\def\gtsima{$\; \buildrel > \over \sim \;$}
\def\simlt{\lower.5ex\hbox{\ltsima}}
\def\simgt{\lower.5ex\hbox{\gtsima}}

\usepackage{color}

\begin{document}
\preprint{ }

\title{Solving differential equations using physics informed deep learning: a hand-on tutorial with benchmark tests}%


\author{Hubert Baty$^1$\footnote{Corresponding author: \href{mailto:hubert.baty@unistra.fr}{hubert.baty@unistra.fr}} , Léo Baty$^{2}$ \\ {\small $^1$Observatoire Astronomique,
 Université de Strasbourg, 67000 Strasbourg, France  \quad \\ $^2$CERMICS, Ecole des Ponts, Marne-la-Vallée,
France}}

\date{\today}%

\begin{abstract}
We revisit the original approach of using deep learning and neural networks to solve differential equations by incorporating the knowledge of the equation. This is done by
adding a dedicated term to the loss function during the optimization procedure in the training process.
The so-called physics-informed neural networks (PINNs) are tested on a variety of academic ordinary differential equations in order to highlight the benefits and drawbacks of
this approach with respect to standard integration methods. We focus on the possibility to use the least possible amount of data into the training process.
The principles of PINNs for solving differential equations by enforcing physical laws via penalizing terms are reviewed. A tutorial on a simple equation model illustrates how
to put into practice the method for ordinary differential equations. Benchmark tests show that a very small amount of training data is sufficient
to predict the solution when the non linearity of the problem is weak. However, this is not the case in
strongly non linear problems where a priori knowledge of training data over some partial or the whole time integration interval is necessary.
\end{abstract}


\maketitle

\clearpage

\section{Introduction}

Neural networks (NN) are widely used to solve problems in a variety of domains including computer vision, language processing, game theory, etc., as one can see in Le Cun et al. (2015) and references therein.
The use of machine learning approaches in the field of scientific computing including differential equations is relatively recent.
Indeed, the idea of leveraging prior knowledge of the physics in the learning process of a NN network was introduced by Raissi et al. (2017, 2019).

Among other things, NN are a tool that can be used for supervised learning, one of the main machine learning settings.
Supervised learning consists in finding a mapping function between given input objects and their associated output values.
This is done by using knowledge about a dataset containing several input/output pairs.
This dataset is used to parameterize the NN such that it minimizes the error between solutions predicted by the NN and true known solutions in the dataset.
The convergence is achieved by minimizing a loss function which expression is based on the mean squared error.
Finding ``good'' parameters is achieved by solving an optimization problem using a gradient algorithm that relies on automatic differentiation to back-propagate gradients through the network (Baydin et al. 2018).

In the case of differential equations, we can apply the supervised learning setting.
Indeed, solving a given differential equation comes down to finding a mapping function between some physical input variable values (position, time, \dots) and a corresponding unknown physical quantity which is the solution of the equation.
Hence, by training a NN we can obtain a non-linear approximation of the solution, which can be used to instantaneously predict the equation's solution at any given input point.
However, a first strong limitation comes from the impossibility to extrapolate the desired solution for input variable values situated outside the range of the training data.
In other words, the NN is a bad extrapolation function.
Second, a minimum amount of training data is required, otherwise, wrong solutions or even absence of convergence can occur during the training process.

In order to tackle these limitations, classical NN can be enhanced by giving it additional information corresponding to the physics.
These approaches are generally called physics-informed neural networks (PINNs), in the context of simulating physical and engineering systems modeled by differential equations.
The method consists in evaluating the solution at some other set of data points (called \emph{collocation points}) at which the estimated solution
must ensure the equations.
A new loss function corresponding to the physics is thus defined and added to the previous one
in the learning process.
In other words, the training is penalized by this additional constraint.
The space of available solutions is thus restricted, being partly driven by the original data and also partly driven by the physics. 
When solving partial differential equations (PDEs), one is particularly interested in using only a very minimal known data set, as for example
the solution on the boundary and at the initial time.
Then, in this sense, the PINNs approach can be said to be mostly physically-driven, as opposed to data-driven.
In this work, we focus on such motivation, even if PINNs can also be used for many other aims like inverse or physics discovery problems (see the discussion and conclusion in the last section).
One can also refer to Cuomo et al. (2022) and Karniadakis et al. (2021) for reviews.

The paper is organized as follows.
We first review the basics of PINNs for PDEs in Section 2.
Section 3 is devoted to a tutorial of the method to solve a simple first order ordinary equation.
The results of benchmark tests performed on a series of different academic ordinary differential equations (with increasingly non-linearity) are presented in Section 4.
Finally, a discussion and conclusions are drawn in Section 5, with a particular attention on highlighting the advantages and drawbacks of the PINN approach versus standard integration schemes.

\section{Physics-Informed Neural Networks}

\subsection{The basics of PINNs for PDE}

We consider a partial differential equation (PDE) written in the following residual form
\begin{equation}
   \mathcal{F} (\boldsymbol{x}, t, u_{\boldsymbol{x}}, u_t, ...) = 0,   \ \ \ \ \ \    \boldsymbol{x} \in \Omega, t \in  \left[ t_0,T \right] ,
\end{equation}
with the imposed initial condition $u(\boldsymbol{x}, t_0) = u_0(\boldsymbol{x})$. A boundary condition must be also specified as,
$u(\boldsymbol{x}, t) = u_\Gamma (t)$ for $\boldsymbol{x} \in  \partial \Omega$ (a Dirichlet-like condition is chosen for simplicity).
$\Omega$ and $\partial \Omega$ represent the spatial domain and associated contour, respectively.
The variables $\boldsymbol{x} \in \mathbb{R}^d$ ($d$ being a spatial dimension) and $t \in \mathbb{R}$ denote the spatial coordinate and time, respectively.
 Note that higher order differential operators can be also included in correspondence with multidimensional system of equations instead of Equation 1.

We introduce a neural network approximating the desired solution $u(\boldsymbol{x}, t)$ of the PDE with $\hat{u_\theta} \simeq u$, where $\theta$ is a set of model parameters.
An artificial feed forward neural network is taken (see Figure 1), with neurons organized in different layers in order to perform calculations
in a sequential way. 
A single input layer containing the input variables $(\boldsymbol{x}, t)$ is connected to a few hidden layers (two layers with four neurons in the schematic example of Figure 1), and finally to an output layer for the solution
$\hat{u_\theta}$. Neurons are only connected in adjacent layers, and are not linked inside each layer. The neural network of $L+1$ layers is a non linear approximation function,
$\mathcal{N}^L (\boldsymbol{z})$, that can be expressed recursively as follows.
The input vector is denoted by $ \boldsymbol{z} \in \mathbb{R}^d_i $, with $d_i = d + 1$, as it includes the spatio-temporal coordinates $(\boldsymbol{x}, t)$.
Thus,
\begin{equation}
 \mathcal{N}^0 ( \boldsymbol{z}) = \boldsymbol{z} .
\end{equation}
For the hidden layers ($1 \le l \le L - 1$), we have
\begin{equation}
 \mathcal{N}^l ( \boldsymbol{z}) =    \sigma ( \boldsymbol{W}^l  \mathcal{N}^{l-1} ( \boldsymbol{z}) +  \boldsymbol{b}^l ) ,
\end{equation}
where we denote the weight matrix and bias vector in the $l$-th layer by $\boldsymbol{W}^l  \in \mathbb{R}^{d_{l-1}  \times d_l}  $  
and $\boldsymbol{b}^l   \in \mathbb{R}^{d_{l} }$ ($d_l$ being the dimension of the input vector for the $l$-th layer). $\sigma(.)$ is a non linear
activation function, which is applied element-wisely. In this work, me choose the most commonly used hyperbolic tangent $tanh$ function.
For the (final) output layer, we get
\begin{equation}
 \mathcal{N}^L ( \boldsymbol{z}) =    \boldsymbol{W}^L  \mathcal{N}^{L-1} ( \boldsymbol{z}) +  \boldsymbol{b}^L ,
\end{equation}
and finally $\hat{u_\theta} (\boldsymbol{x}, t) = \mathcal{N}^L ( \boldsymbol{z})$. The network can be also written as a sequence of non linear functions
\begin{equation}
\hat{u_\theta} (\boldsymbol{x}, t) =  ( \mathcal{N}^L \circ \mathcal{N}^{L-1} ...\  \mathcal{N}^0) (\boldsymbol{x}, t) ,
\end{equation}
where the operator $\circ$ denotes the composition and $\theta =  \lbrace \boldsymbol{W}^l,  \boldsymbol{b}^l  \rbrace_{l=1,L}$ represents the trainable parameters (weight matrices and bias vectors) of the network.
The goal is to calibrate its parameters $\theta$ such that $\hat{u_\theta} $ approximates the target solution $u(\boldsymbol{x}, t)$.
 
  \begin{figure}[!t]
\centering
 \includegraphics[scale=0.36]{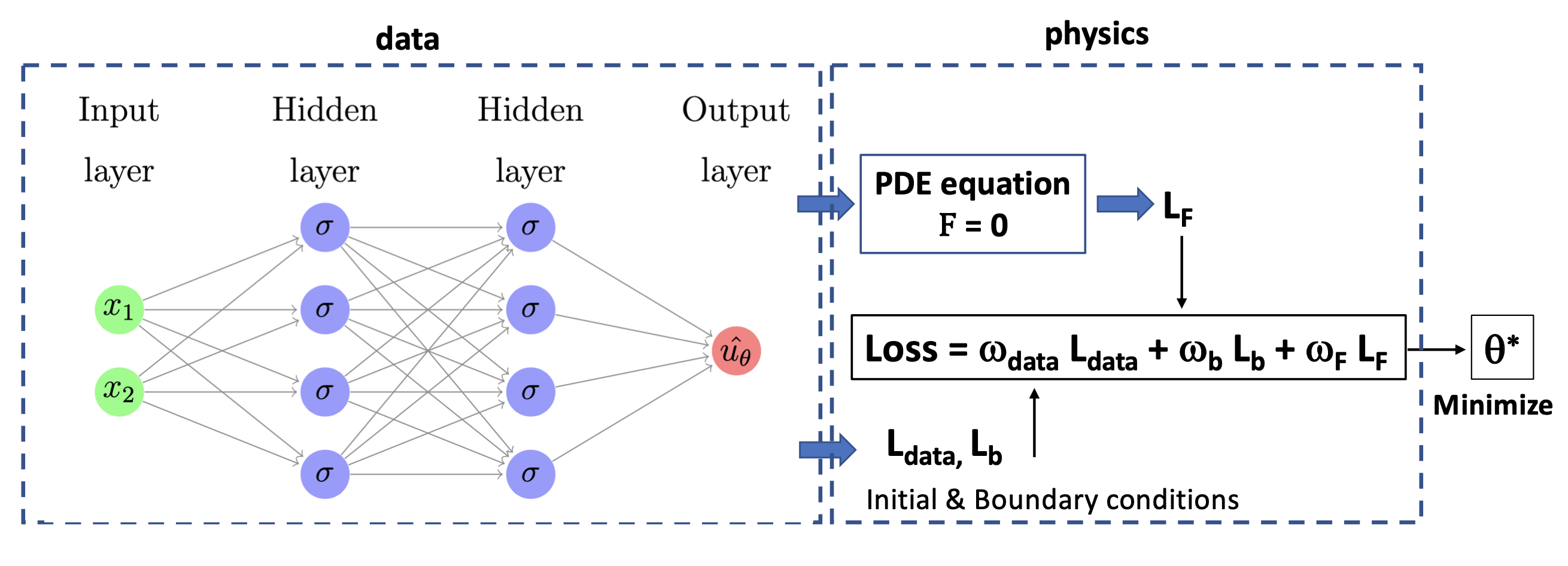}
  \caption{Schematic representation of the structure for a Physics-Informed Neural Network applied to the resolution of a differential equation. The input layer has two input variables (i.e. two neurons) noted $x_1$ and $x_2$
  representing for example a space and a time coordinate respectively. Two hidden layers with four neurons per layer are connected with the input and the output layer, where the latter has a single variable (one neuron) representing the desired solution
  $\hat{u_\theta}$.
   }
\label{fig1}
\end{figure}

\subsection{The training of PINNs for PDE}

The resolution of the PDE is reduced to an optimization problem as follows, and as schematized in Figure 1. We first assume that a set of $N_{data}$ data
is available for the known solution at different times, i.e. 
$\lbrace t_{data}^i, \boldsymbol{x}_{data}^i, {u}_{data}^i \rbrace _{i=1}^{N_{data}}$ that are the training data, which include
the initial condition. A corresponding loss function $L_{data}$ (using the mean square error formulation) can be deduced from the residual as
\begin{equation}
   L_{data} (\theta) = \frac  {1} {N_{data} } \sum_{i=1}^{N_{data} } \left\|\ \hat{u_\theta} (\boldsymbol{z_i} ) - u_{data}^i  \right\|^2 .
\end{equation}
In a similar way, defining a loss function $L_{b}$ corresponding to the knowledge of the boundary condition, we have
\begin{equation}
   L_{b} (\theta) = \frac  {1} {N_b} \sum_{i=1}^{N_b} \left\|\ \hat{u_\theta} (\boldsymbol{z_i} ) - u_b^i  \right\|^2 ,
\end{equation}
where a set of $N_b$ known data is imposed via $\lbrace t_{b}^i , \boldsymbol{x}_{b}^i, {u}_{b}^i \rbrace _{i=1}^{N_{b}}$.
Finally, another loss function for the equation itself can be also obtained as,
\begin{equation}
   L_{ \mathcal{F}} (\theta) = \frac  {1} {N_c} \sum_{i=1}^{N_c}  \left\|\  \mathcal{F} ( \hat{u_\theta} (\boldsymbol{z_i} )  )   \right\|^2 ,
\end{equation}
that must be evaluated on a set of $N_c$ data points (generally called collocation points) as explained below.
Indeed, one advantage of the neural network approach is given by the possibility to evaluate exactly the differential operators at the collocation points in  $L_{ \mathcal{F}}$
and $\mathcal{F}$ by using automatic differentiation. The automated differentiation is also used to compute derivatives with respect to the network weights (i.e. $\theta$),
that is necessary to implement the optimization procedure (see below). Note that in this way, the derivatives can be obtained at machine precision, contrary to the use of
some standard numerical scheme. Moreover, the latter operations are greatly facilitated by Python open source software libraries like Tensorflow or Pytorch.

A composite loss function is generally formed as
\begin{equation}
              L  (\theta)    =   \omega_{data} L_{data} (\theta)  + \omega_{b} L_{b} (\theta) + \omega_{\mathcal{F}}L_{ \mathcal{F}} (\theta),
\end{equation}
where an optimal choice of values for hyper-parameters $(\omega_{data}, \omega_{b},  \omega_{\mathcal{F}})$ allow to ameliorate the eventual unbalance between
the partial losses during the training process. These weights can be user-specified or automatically tuned. In the present work, for simplicity we fix the $\omega_{data} $
value
to be constant and equal to unity, and the other weight parameters are determined with values varying from case to case.

A gradient descent algorithm is used until convergence towards the minimum is obtained for a predefined accuracy (or a given maximum iteration number) as
\begin{equation}
             \theta^{i+1} =  \theta^{i} - \eta  \nabla_{\theta}     L  (\theta^i) ,
\end{equation}
for the $i$-th iteration also called epoch in the literature,
leading to $  \theta^{*}  = \operatorname*{argmin}_\theta  L  (\theta)$, where $\eta$ is known as the learning rate parameter.
In this work, we choose the well known $Adam$ optimizer. A standard automatic differentiation technique is necessary to compute derivatives (i.e. $\nabla_{\theta}$) with respect to
the NN parameters (e.g. weights and biases) of the model (Raissi et el. 2019).

\subsection{The ODE case}

In this study, we focus on ordinary differential equations (ODE). Thus, the spatial dependence is ignored in Equation 1, and we are left with a desired solution
$u(t)$ and its approximation $ \hat{u_\theta} (t)$. The input first layer of the neural network is supplied with $N_{data}$ values at different times $t_i \in  \left[ t_0,T \right] $
corresponding to $u_{data} (t_i) = u_{data}^i$ (with $i = 1, N_{data}$). The initial condition corresponds to the first point,  $u_{data} (t_0) = u_{data}^1$. In this work, for simplicity, we
assume a uniform distribution of the $N_{data}$ points within a subinterval of the full time interval domain. As a consequence of the ODE particular case, we are not concerned
with the boundary condition, thus $\omega_b = 0$. However, the central part of the PINNs concept concerns the loss function $ L_{ \mathcal{F}} (\theta)$, that is evaluated at $N_c$
collocation points which are not necessarily coinciding in time values with $N_{data}$. The distribution of $N_c$ is also taken to be uniform in the full time domain, or 
on a subinterval of it.

Some of the Pytorch Python-based codes and data-sets accompanying this manuscript are available on the GitHub repository at https://github.com/hubertbaty/PINNS-EDO. These have been inspired
by the codes provided on https://benmoseley.blog and available on GitHub repository at https://github.com/benmoseley/harmonic-oscillator-pinn.
We have chosen to use very simple deep
feed-forward networks architectures with hyperbolic tangent activation functions. In this work, the optimal choice of detailed architecture of the network (number of hidden layers, 
number of neurons per layer) and of hyperparameters (learning rate, loss weights) is done manually. Although more systematic/automatic procedures could be used,
this is a more complicated task not considered in this work.

Note that we use the notation $y$ for the desired ODE solution in the following instead of $u$ introduced in the previous section for a PDE case.

\begin{figure}[!ht]
\centering
 \includegraphics[scale=0.45]{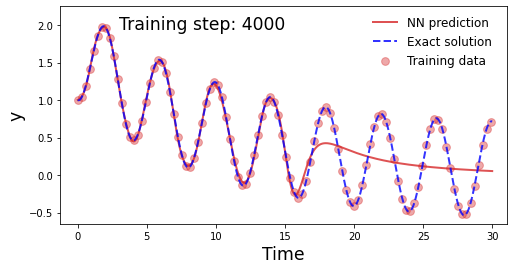}
 \includegraphics[scale=0.45]{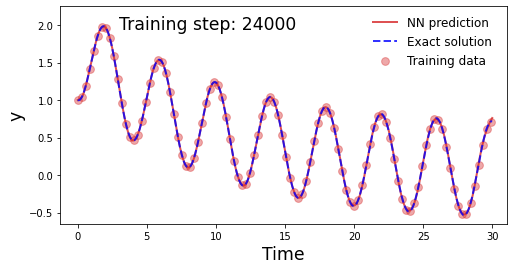}
  \caption{Tutorial solution $ \hat{y_\theta} (t)$ (red solid line) predicted by the normal NN for $n_t = 4000$ iterations (in left panel) and $n_t = 24000$ iterations (in right panel)
  respectively, and compared to the exact solution (blue hatched line). The chosen training data set values
  are indicated using circles (with $N_{data} = 101$). The physical information is not used, i.e. $\omega_{ \mathcal{F}} = 0$.
   }
\label{fig2}
\end{figure}

\begin{figure}[!ht]
\centering
 \includegraphics[scale=0.5]{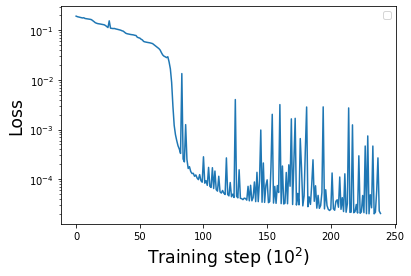}
  \caption{Loss function as a function of the number of epochs (i.e. number of iterations $n_t$), corresponding to the previous figure.
   }
\label{fig3}
\end{figure}

\section{Illustration of the method on a simple tutorial example}
\label{setup}

\subsection{The differential equation: a tutorial example}
Let us consider the following equation example, called tutorial equation below, to be solved for $y(t)$,
\begin{equation}
         \frac {dy} {dt} + 0.1t - \sin (\pi t/2) = 0,
 \end{equation}
fo $t \in \left[ 0,30 \right] $, using the initial condition $y_0 = y(0) = 1$. As can be seen below, the corresponding solution contains two time scales, a first one
due to the sinusoidal forcing term, and a second one due to the linear term in $\frac {t} {10}$ that gives an exponentially decreasing envelope amplitude.

\begin{figure}[!t]
\centering
 \includegraphics[scale=0.45]{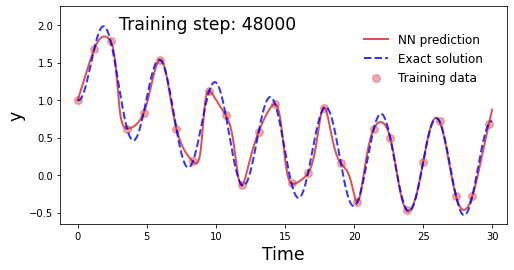}
 \includegraphics[scale=0.45]{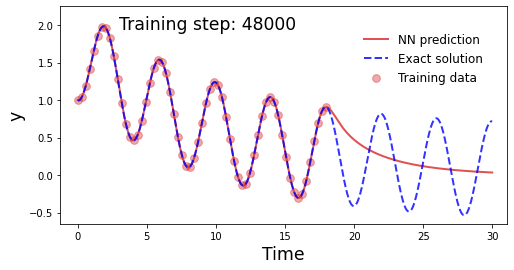}
  \caption{Tutorial solution obtained in two cases using the normal NN, for a low number of data ($N_{data} = 26$) uniformly taken within the whole time domain (left panel), and for
   $N_{data} = 61$ taken within a left subinterval (right panel). The convergence is stopped at $n_t = 48000$.
   }
\label{fig4}
\end{figure}

\subsection{Solving with a normal neural network}
We first consider a situation without any constraint coming from the differential equation, i.e. $\omega_{ \mathcal{F}} = 0$. As we are left
with data coming only from the exact solution, we call it the normal neural network. The training procedure is illustrated on
Figure 2  for two training steps (i.e. $n_t$), at $n_t = 4000$ and $n_t = 24000$. The solution is not fully converged for $n_t = 4000$, contrary
to $n_t = 24000$. This is in agreement with the history of the loss function (see Figure 3), where the convergence is already roughly obtained when
$n_t  \simeq 15000$.
A learning rate of $ \eta = 3  \times 10^{-3}$, with $ \omega_{data} = 1$ is chosen.  A choice of $3$ hidden layers with $32$ neurons per layer is also done.
The exact solution that is drawn for comparison, is obtained using a classical Runge-Kutta method of order two. Note that the later is also useful
to extract the training data corresponding to the $N_{data}$ values, $y_{data}^i $.  Figure 2 clearly illustrates the ability of the normal NN to
approximate the solution for a relatively high number of $N_{data}$ values, as $N_{data} = 101$ is employed.

However, when the number of training points is not enough, the convergence towards the solution is bad, or it can completely fail.
This is illustrated in two cases at the end of the convergence process stopped at $n_t = 48000$.
Indeed, the results for the first case obtained for $N_{data} = 26$ values uniformly chosen within the full time interval, show bad convergence properties as
illustrated in Figure 4 (left panel). Moreover, when the points are distributed only within a subinterval (with $N_{data} = 61$), Figure 4 (right panel) shows that the
method completely fails to obtain an acceptable solution in the other subinterval that is free of training data.
In other words, the NN is not able to correctly extrapolate the solution in these two cases.

\begin{figure}[!t]
\centering
 \includegraphics[scale=0.42]{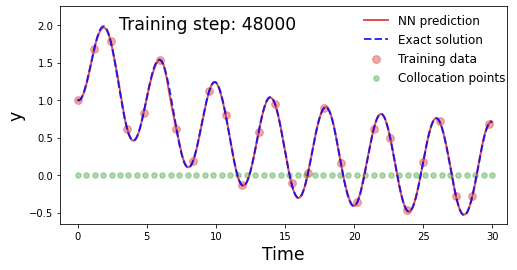}
 \includegraphics[scale=0.42]{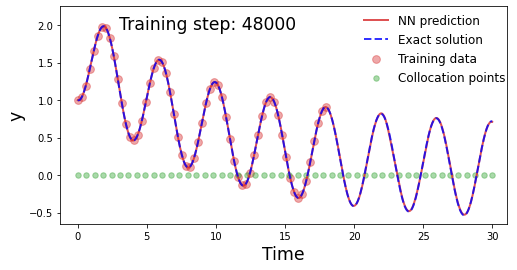}
  \caption{Tutorial solution obtained in the two cases of the previous figure with PINNs, where a set of $N_c = 50$ collocation points is used to calculate an ODE loss function $L_{ \mathcal{F}}$.
  The cases with $N_{data} = 26$ data points taken within the whole interval, and with $N_{data} = 61$ data values within a left subinterval, are plotted in left/right panel respectively. The time values (for collocation points) at which the
  physical loss function is evaluated are indicated with the small green circle on $t$ axis.
   }
\label{fig5}
\end{figure}

\begin{figure}[!t]
\centering
 \includegraphics[scale=0.42]{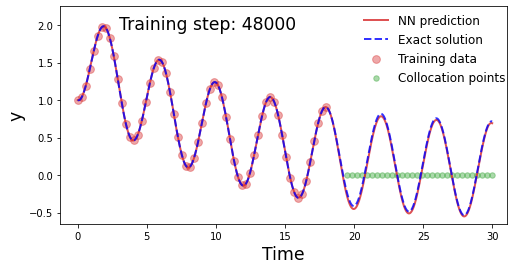}
  \caption{Tutorial solution obtained with PINNs for $N_{data} = 61$ training data values taken within a left subinterval, and $N_c = 30$ collocation points taken within a complementary right subinterval.
   }
\label{fig6}
\end{figure}

\begin{figure}[!t]
\centering
 \includegraphics[scale=0.45]{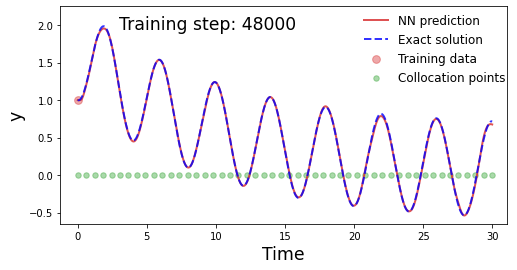}
 \includegraphics[scale=0.45]{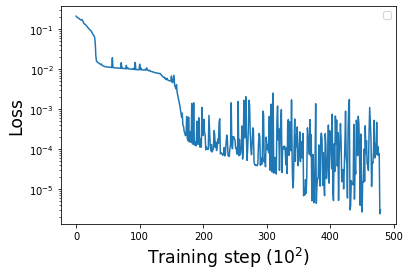}
  \caption{(Left panel) Tutorial solution obtained with PINNs, with $N_{data} = 1$ (initial condition imposed at $t = 0$), and  $N_c = 50$ collocation points. (Right panel) Corresponding history
  of the loss function.
   }
\label{fig7}
\end{figure}

\subsection{Solving with PINNs}
In this sub-section, we consider now the possibility to add the constraint on the loss function with a non zero contribution coming from the differential equation at some chosen collocation points,
i.e. $\omega_{ \mathcal{F}} \neq 0$.
In other words, we minimize by adding a weighted partial loss function term $ \omega_{ \mathcal{F}} L_{ \mathcal{F}} (\theta)$, where $\mathcal{F} = \frac {dy} {dt} + 0.1t - \sin (\pi t/2)$.
Typically, we define a uniform data set of $N_c = 50$ points within the full time interval. The results of the training process is illustrated for the two cases previously studied
using the normal network, for which the convergence is not satisfaying. The results plotted in left/right panels of Figures 5, display a spectacular amelioration after 
$n_t = 40000$ iteration steps in both cases. Additionally, Figure 6 shows that taking collocation points only within a right subinterval (for the second above case) can also be
sufficient.
Note that, a minimum value for $N_c$ is required with an exact value that depends on the parameters of the network (i.e. number of layers, neurons, etc.). The distribution of the collocation points can
also influence the results, but the philosophy behind the PINNs technique remains. The learning rate together with the loss weight values can also influence the convergence of the gradient descent algorithm.
Indeed, a too high value of $\eta$ leads to strong oscillations in the loss function, whilst a too small value can induce a very small convergence speed.
The combination of relative weights ($\omega_{data}$ and $\omega_{\mathcal{F}}$) is also important, in order that the two partial losses converge at
a similar rate.
For the example studied above, we have taken optimal values $\omega_{data} = 1$, and $\omega_{\mathcal{F}} = 6  \times 10^{-2}$ for the weights of the two partial losses. 

If we reduce the training data amount to the minimum possible, i.e. only one point corresponding to the initial value is taken, the solution obtained is also
excellent as one can see in left panel of Figure 7. The corresponding loss function is also plotted in right panel of Figure 7. Note that, in this case, an optimal choice of $4$ hidden layers with $32$ neurons per layer is done.

Now the question is, does the method work so nicely for any (ODE) differential equation for which a very reduced amount of data is known.
In order to answer this question, we investigate a rather large number of academic cases in the following section, including the important class of
second order differential equations.

\section{Different examples - Benchmark tests}
\label{setup}

\subsection{Harmonic oscillator}
In this sub-section, we first consider the following harmonic oscillator equation,
\begin{equation}
         \frac {d^2y} {dt^2} + \omega_0^2 y =0,
 \end{equation}
where $ \omega_0$ is the normalized angular frequency, and where the time domain considered is $t \in \left[ 0,1 \right] $. We also take the initial
conditions $y(t = 0) = y_0 = 1$ and $\frac {dy} {dt}  ( t = 0) = 0$. Thus, the exact solution is a simple $cosine$ function, i.e. $y (t) =  \cos (\omega_0 t)$.
Our PINNs algorithm is first used to integrate the oscillator equation for $\omega_0 = 20$, i.e. for a time interval slightly larger than 3 periods.
Note that, in order to evaluate the corresponding ODE loss function $L_{ \mathcal{F}} $, a second order automatic differentiation must be used for this example.
The choice of the activation function is important to this respect (i.e. the hyperbolic tangent in this study).
The solution predicted when only one point corresponding to the initial value $(N_{data} = 1)$ is imposed, is plotted in left panel of Figure 8. It is clearly
bad for the parameters taken in this case. The quality of the result can slightly varies with this choice, but taking only one training data value is in general
not sufficient to lead to an acceptable solution. This is not surprising, as a classical integration method (either analytical or numerical) requires two initial conditions
for such second order differential equation.

The quality of the solution is greatly ameliorated by adding a second data value (i.e. $N_{data} = 2$). Another second improvement comes from adding
another constraint, that is the conservation of the total energy $E =  \frac {1} {2} ( \frac {dy} {dt})^2 + \frac {1} {2} \omega_0^2 y^2$
(up to addition by a constant which value is determined by the initial conditions). Indeed, we can add to the total loss another partial loss function defined as
\begin{equation}
  L_E (\theta) = \frac  {1} {N_c} \sum_{i=1}^{N_c}  \left\|\  E ( \hat{u_\theta} (\boldsymbol{z_i} ) )- E_0   \right\|^2 ,
 \end{equation}
with a corresponding weight $\omega_E$ to be determined and $E_0$ being the initial constant total energy. More explicitly, we have $L  (\theta)    =   \omega_{data} L_{data} (\theta)  + \omega_{E} L_{E} (\theta) + \omega_{\mathcal{F}}L_{ \mathcal{F}} (\theta)$.
These new results are visible in right panel of Figure 8 for the predicted solution obtained at $n_t = 54000$.
Moreover in Figure 9, one can compare the corresponding mean square error ($MSE$) and loss histories evaluated for $1000$ points taken within the full time domain.
One can clearly see the improvement with a lower minimum $MSE$ value by more than two orders of magnitude obtained in the improved procedure. This is remarkable, as this is despite the fact that
the loss function history (also visible on the same figure) converges in a similar way in the two cases.

\begin{figure}[!t]
\centering
 \includegraphics[scale=0.42]{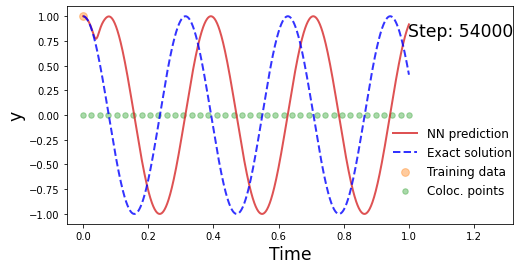}
 \includegraphics[scale=0.42]{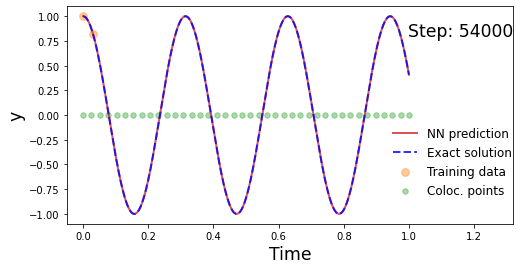}
  \caption{(Left panel) Harmonic oscillator predicted solution obtained with PINNs, using $N_{data} = 1$ (initial condition imposed at $t = 0$), and  $N_c = 40$ collocation points. (Right panel)
  Predicted solution obtained with PINN, with $N_{data} = 2$ and the additional constraint on the energy conservation (see text).
   }
\label{fig8}
\end{figure}

\begin{figure}[!t]
\centering
 \includegraphics[scale=0.51]{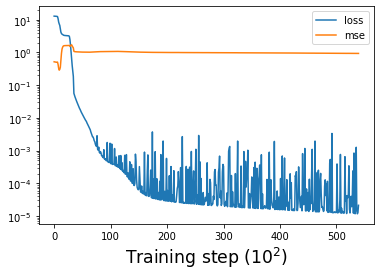}
 \includegraphics[scale=0.51]{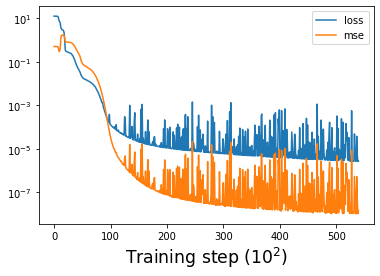}
  \caption{Histories of the total loss function $L  (\theta)$ and $MSE$ corresponding to the two panels of the previous figure respectively. The $MSE$ is evaluated using the standard
  expression, $MSE =   \frac  {1} {N_{eval} } \sum_{i=1}^{N_{eval} } \left\|\ \hat{y_\theta} (t_i) - y_{eval}^i  \right\|^2 $, where the evaluation $\hat{y_\theta} (t_i)$ is done
  on $N_{eval} = 1000$ points uniformly distributed within the whole time interval, and $y_{eval}^i$ is the expected exact solution at $t=t_i$.
   }
\label{fig9}
\end{figure}

Note also that, with these two improvements we have obtained that a relatively low minimum value for the number of collocation points with $N_c \simeq 24$ is sufficient.
Such very small minimum number of collocation points is a great advantage compared to a standard integration method for which the small time step restriction (because of stability and/or
precision) requires a much larger number of points within the time interval.
The choice of the other hyperparameters are $ \eta = 3  \times 10^{-4}$, $\omega_{data} = 1$, and $\omega_{\mathcal{F}} = \omega_E = 3  \times 10^{-4}$.
Three hidden layers with $32$ neurons per layer are taken for the neural network architecture. We have also investigated longer time cases with higher $\omega_0$ values, typically
up to $\omega_0 \simeq 130$ (not shown). The conservation of energy is fundamental to this respect, otherwise the algorithm doesn't converge towards the expected solution for such long
time integration.

\subsection{Non linear pendulum}
Second, as a natural extension of the harmonic oscillator, we consider the non linear pendulum example below,
\begin{equation}
         \frac {d^2y} {dt^2} + \omega_0^2 \sin(y) =0,
 \end{equation}
where $ \omega_0$ is the normalized angular frequency (a value $ \omega_0 = 25$  is taken below), and where the time domain considered is $t \in \left[ 0,1 \right] $. We also choose the initial
conditions $y_0 = 0.1$ and $\frac {dy} {dt}  ( t = 0) = 40$. As for the harmonic oscillator, we investigate the possibility to use one single point
for $N_{data}$  corresponding to the initial condition $y_0$, the conservation of the total energy $E$ being imposed via a non zero $ \omega_{E} L_{E} (\theta)$ term with
$E = \frac {1} {2} ( \frac {dy} {dt})^2 - \omega_0^2 \cos (y)$ (up to addition by a constant which value is determined by the initial conditions).
The hyperparameters used are  $\eta = 1  \times 10^{-3}$, $\omega_{data} = 1$, $\omega_{\mathcal{F}} = 3  \times 10^{-6}$, and
$\omega_E = 3  \times 10^{-7}$. When the NN architecture is composed of $4$ hidden layers with $32$ neurons per layer, the converged solution
obtained for $72000$ epochs is correct. However, taking $3$ layers (instead of $4$) leads to a wrong solution even if the loss function
displays convergence. Indeed, in the latter case, the calculated solution is shifted with respect to the exact solution, and the $MSE$
is dramatically high. This is illustrated in Figures 10 and 11.

\begin{figure}[!t]
\centering
 \includegraphics[scale=0.42]{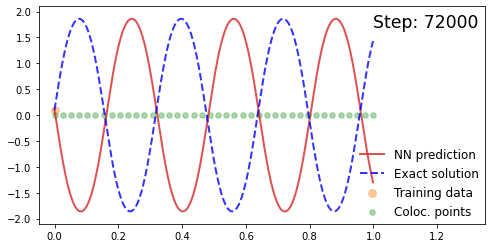}
 \includegraphics[scale=0.42]{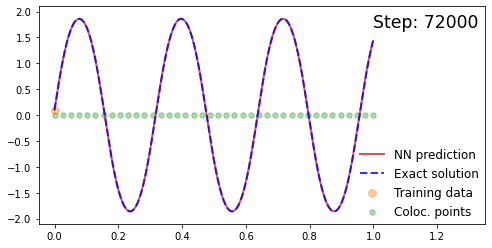}
  \caption{PINNs solution obtained for the non linear pendulum with $N_{data} = 1$ and  $N_c = 40$ collocation points, using a NN architecture of
  $3$ hidden layers (left panel) and $4$ hidden layers (right panel).
   }
\label{fig10}
\end{figure}

\begin{figure}[!t]
\centering
 \includegraphics[scale=0.5]{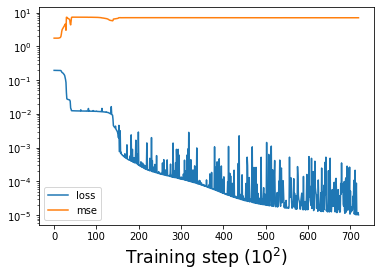}
 \includegraphics[scale=0.5]{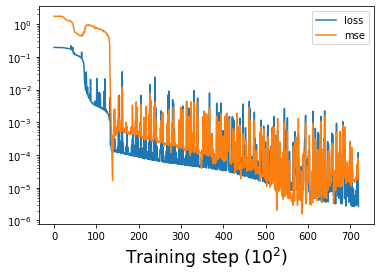}
  \caption{History of the total loss function  $L  (\theta)$ and $MSE$ corresponding to the two cases of the previous figure (left and right panel for  $N_{data} = 1$ and $N_{data} = 2$
  respectively). The $MSE$ is evaluated using the standard
  expression, $MSE =   \frac  {1} {N_{eval} } \sum_{i=1}^{N_{eval} } \left\|\ \hat{y_\theta} (t_i) - y_{eval}^i  \right\|^2 $, where the evaluation $\hat{y_\theta} (t_i)$ is done
  on $N_{eval} = 1000$ points uniformly distributed within the whole time interval, and $y_{eval}^i$ is the expected exact solution at $t=t_i$.
   }
\label{fig11}
\end{figure}

Of course, taking two data points (i.e. $N_{data} = 2$) allows a nice convergence towards the expected exact solution (not shown) for $3$ and $4$ hidden layers.
And, as for the harmonic oscillator equation, the minimum required value for the number of colocation points $N_c$ remains rather low, as it is of order $35$ now.

We have also investigated the possibility to use an equivalent form of a system of two first order differential equations, as done in analytical or classical
numerical integrations. Indeed, we can consider the following equivalent system:

\begin{equation}
\centering
\left\{\begin{split}
 \frac {dy_1} {dt} - \omega_0 y_2 =0, \\
 \frac {dy_2} {dt} + \omega_0 \sin(y_1) = 0,
\end{split}\right.
 \end{equation}
where $y_1$ represents the desired solution (i.e. the previous $y$ parameter) and $y_2$ is its associated time derivative $\frac {dy} {dt}$ divided by $\omega_0$.
The advantage of the latter normalisation is important, as it facilitates the use of the NN network because in this way $y_1$ and $y_2$ have values
of the same order of magnitude. Otherwise, two networks (one per variable) must probably be employed.
We have thus used our PINNs algorithm with an input layer containing one neuron for $t$, $4$ hidden layers with $32$ neurons per layer, and a final
output layer containing two neurons for $y_1$ and $y_2$.  The partial loss function for the data, $L_{data}$, is calculated using two points at $t = 0$ now, one 
value for $y_1 (t = 0) = 0.1$ and one for $y_2 (t = 0) = 1.6$. The procedure is equivalent to a standard integration scheme using one initial condition for the solution and one
for its derivative. The partial loss for the ODE, $ L_{ \mathcal{F}} = L_{ \mathcal{F}_1}  + L_{ \mathcal{F}_2} $, is now the sum of two terms corresponding
to the two equations respectively. The results of the previous example with $N_c = 40$ are successfully obtained using the following hyper parameters,
$\eta = 3  \times 10^{-3}$, $\omega_{data} = 1$, $\omega_{\mathcal{F}} = 1  \times 10^{-1}$, and $\omega_E =  2  \times 10^{-6}$, as plotted
in Figure 12. The corresponding $MSE$ is similar to the $MSE$ previously obtained for the PINN solution of the single second order equation using $N_{data} = 2$, with the constraint on the
energy conservation via a $L_E$ partial loss function. Note that, the total energy with the system formulation is, $E = \omega_0^2 \left ( \frac {y_2^2} {2} - \cos(y_1) \right )$ (up to a addition by a constant).

\begin{figure}[!t]
\centering
 \includegraphics[scale=0.5]{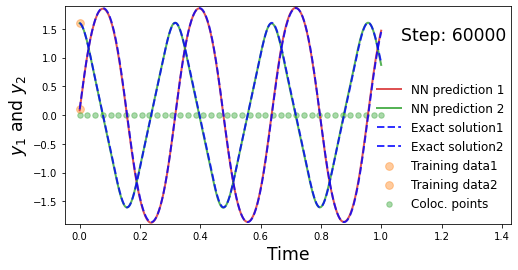}
 \includegraphics[scale=0.5]{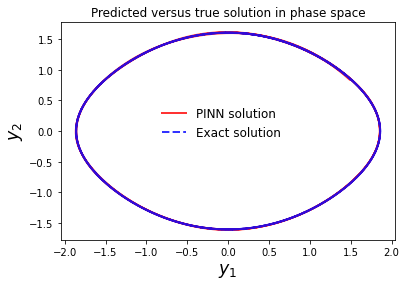}
  \caption{PINNs solution obtained for the non linear pendulum using an equivalent formulation with two equations for the two variables $y_1$ and $y_2$ (see text) with $N_{data} = 2$ (one point for $y_1$ and one for $y_2$, i.e.
  for the two initial conditions at $t = 0$) and  $N_c = 40$ collocation points. The energy conservation is also applied. Solutions as functions of time and in phase space are plotted
  in left and right panels, respectively.}
\label{fig12}
\end{figure}

\subsection{Anharmonic oscillators}
If we consider an anharmonic potentiel of the form $\frac {y^4} {4} $ (instead of $\frac {y^2} {2}$ for the harmonic oscillator), we get the corresponding ODE with a non linear restoring force
$\propto y^3$,
\begin{equation}
         \frac {d^2y} {dt^2} + \omega_0^2  {y^3} = 0.
 \end{equation}
The PINNs integration for this problem leads to results and conclusions very similar to the non linear pendulum case. Indeed,
the use of two training data points with energy conservation constraint considerably ameliorate the convergence. Thus we have $E = \frac {1} {2} ( \frac {dy} {dt})^2 + \frac {1} {4} \omega_0^2 y^4$
(up to addition by a constant which value is determined by the initial conditions).
This is illustrated in Figure 13 for a case with $y_0 = 1.5$ and $\omega_0 = 15.5$.
We have chosen the following hyper parameters, $\eta = 1.5  \times 10^{-3}$, $\omega_{data} = 1$, $\omega_{\mathcal{F}} = 1  \times 10^{-5}$, and $\omega_E =  1  \times 10^{-6}$.
The choice of five hidden layers for the neural network seems to be a better optimal choice for this equation.
Note also that for the latter example, a minimum number of collocation points of $N_c = 37$ seems to be necessary, that is slightly higher than for the non linear pendulum. However, as
we have not investigated a large range of different initial parameters, this conclusion is not firm and only gives a rough tendency.

\begin{figure}[!t]
\centering
 \includegraphics[scale=0.5]{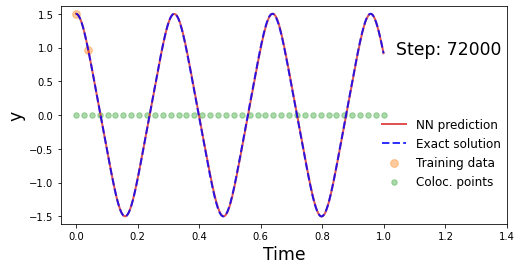}
  \caption{PINN solution obtained for the anharmonic oscillator with $N_{data} = 2$ and $N_c= 40$.}
\label{fig13}
\end{figure}

It is also instructive to consider a more complex dynamical system corresponding to a double well potential of the form $\frac {y^4} {4} - \frac {y^2} {2}$,  and according to the differential equation,
\begin{equation}
         \frac {d^2y} {dt^2} + \omega_0^2   ( y^3 -  y ) = 0.
 \end{equation}
For this example, two families of solution exist (see below). In order to easier impose the initial conditions, we also take the equivalent system of two equations (as done
for the non linear pendulum),
\begin{equation}
\centering
\left\{\begin{split}
 \frac {dy_1} {dt} - \omega_0 y_2 =0, \\
 \frac {dy_2} {dt} + \omega_0 ( y_1^3 - y_1 ) = 0.
\end{split}\right.
 \end{equation}
Indeed, taking the initial condition $y(0) = y_0 = 1.8$ , or equivalently $y_1 (0) = 1.8$, together with zero derivative $y_2 (0) = 0$, a first solution is obtained
as plotted in Figure 14. We use $\omega_0 = 12$. The predicted solution is nicely reproduced when compared to the exact expected solution (obtained using a Runge-Kutta integration).
For this example, two training data points seems to be necessary (i.e. $N_{data} = 4$), with the use of the energy conservation as $E = \omega_0^2 ( \frac {y_2^2} {2} +  \frac {y_1^4} {4}  -  \frac {y_1^2} {2} )$.
We have also chosen the following hyper parameters, $\eta = 1  \times 10^{-3}$, $\omega_{data} = 1$, $\omega_{\mathcal{F}} = 6  \times 10^{-3}$, and $\omega_E =  6  \times 10^{-5}$, 
but they must be adjusted from case to case.
A second solution corresponding to $y(0) = y_0 = 1.38$ (with zero initial derivative condition) is also trained and nicely obtained, as one can see in Figure 15.

\begin{figure}[!t]
\centering
 \includegraphics[scale=0.45]{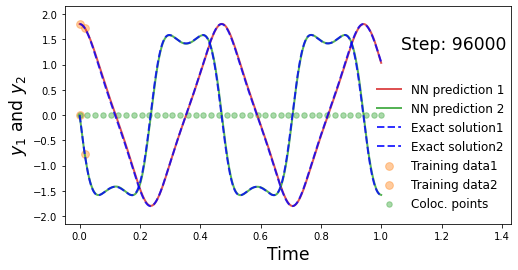}
 \includegraphics[scale=0.45]{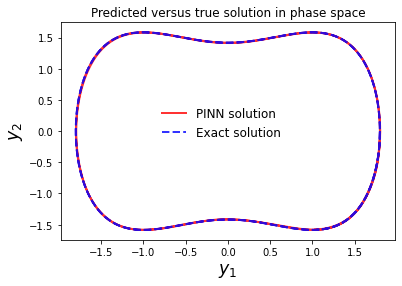}
  \caption{PINNs solution obtained for the double well oscillator system for the initial condition $y_0 = 1.8$. We use $N_{data} = 4$ with $2$ points for each variable,  and $N_c = 40$ collocation points.
  A NN architecture with five hidden layers is taken.
   The energy conservation is also applied. Solutions are plotted as functions of time (left panel), and in phase space (right panel).}
\label{fig14}
\end{figure}

\begin{figure}[!t]
\centering
 \includegraphics[scale=0.45]{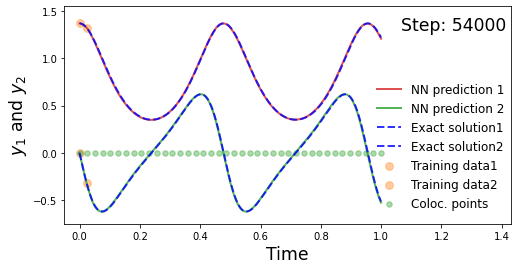}
 \includegraphics[scale=0.45]{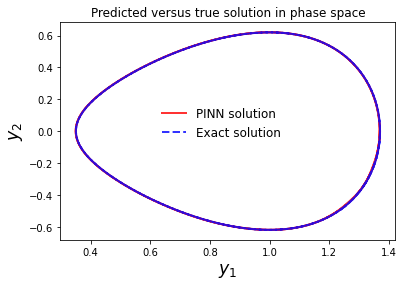}
  \caption{PINNs solution obtained for the double well oscillator system for the initial condition $y_0 = 1.38$. We use $N_{data} = 4$ with $2$ points for each variable,  and $N_c = 40$ collocation points.
  A NN architecture with five hidden layers is taken.
   The energy conservation is also applied. Solutions are plotted as functions of time (left panel), and in phase space (right panel).}
\label{fig15}
\end{figure}

However, taking an initial condition closer to the critical value $y_0 = \sqrt 2$, our PINN algorithm fails to converge to the exact solution. This is not completely
surprising, as this corresponds to a threshold separating solutions having orbits in the phase space trapped in the well centered on ($ y_1 = 0 , y_2 = 0$) with
solutions having orbits centered on ($ y_1 = 1 , y_2 = 0$). In other words, two types of solutions coexist for  $y_0 = \sqrt 2$.

\subsection{Van Der Pol oscillator}

We now consider the Van Der Pol (VDP) oscillator equation given by,
\begin{equation}
         \frac {d^2y} {dt^2} + \omega_0^2  {y} - \epsilon   \omega_0 (1 - y^2)  \frac {dy} {dt} = 0,
 \end{equation}
where $ \omega_0$ is a normalized angular velocity, and $\epsilon$ is a parameter having a value which determines the amplitude of a limit cycle in the phase space (see below).
Note that the harmonic oscillator is recovered for $\epsilon = 0$.
The particularity of the system is the existence of a limit cycle as illustrated in Figure 16, obtained using a classical Runge-Kutta integration (of order $4$) for
 $ \omega_0 = 15$, and $\epsilon = 5$, for $t \in \left[ 0,3 \right] $. 
When the parameter $\epsilon$ has a lower value, the system exhibits a limit cycle with less distortion in the phase space, as illustrated in Figure 17.

\begin{figure}[!t]
\centering
 \includegraphics[scale=0.49]{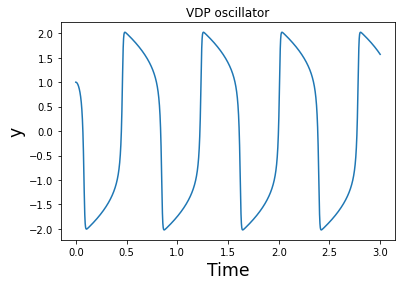}
 \includegraphics[scale=0.47]{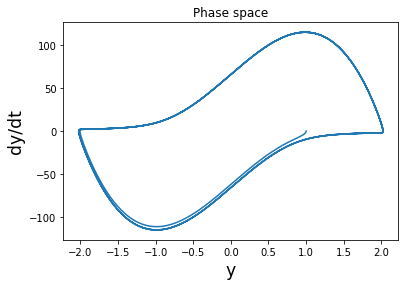}
  \caption{Exact solution of the VDP oscillator equation obtained for $\epsilon = 5$ and the initial conditions $y (t = 0) = 1$, $ \frac {dy} {dt}  (t = 0) = 0$. The solution $y(t)$ as a function of
  time is plotted in the left panel, and in phase space in the right panel.}
\label{fig16}
\end{figure}

\begin{figure}[!t]
\centering
 \includegraphics[scale=0.49]{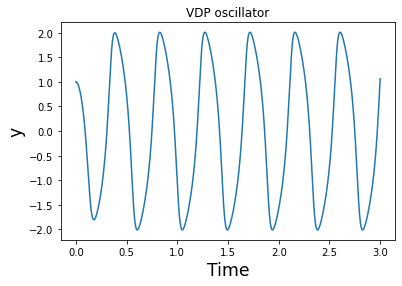}
 \includegraphics[scale=0.47]{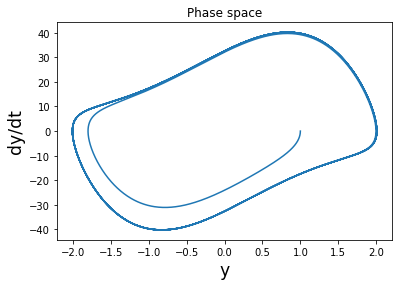}
  \caption{Same as in the previous figure for $\epsilon = 1$.}
  \label{fig17}
\end{figure}

We have thus used our PINNs algorithm to explore its behavior for three cases corresponding to three values of $\epsilon$, i.e.  $\epsilon =  \frac {1} {3} ,1$, and $5$.
We have chosen $ \omega_0 = 15$ for $t \in \left[ 0,1.5 \right] $. 
We also take the initial condition $y_0 = 1$. The results are plotted in Figure 18, for a neural network having three hidden layers with $32$
neurons per layer. The number of training data and collocation points, as well as the weight $\omega_{\mathcal{F}}$ associated to the equation ($\omega_{data} = 1$ being fixed) are varying from case to case (see the legend).
The results clearly show that increasing the non linearity (via the $\epsilon$ parameter) require a higher number of points. This is the case of the number of
collocation points, but also of the number of training data points. Indeed, for the $\epsilon = 5$ case we need a collection of training data points distributed within the whole time domain,
while for the smallest $\epsilon$ case, three points at early times are sufficient to obtain a convergence towards the exact solution. This is clearly a strong
limitation of the PINN algorithm when one want to solve highly non linear problems.

\begin{figure}[!t]
\centering
 \includegraphics[scale=0.5]{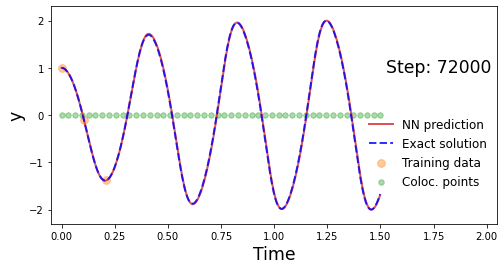}
 \includegraphics[scale=0.5]{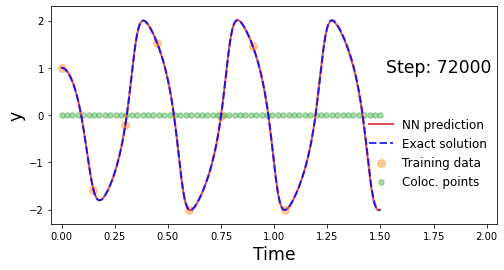}
 \includegraphics[scale=0.5]{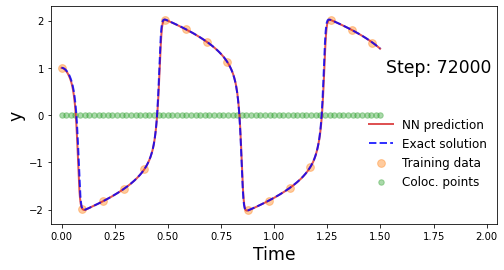}
  \caption{PINN solution of the VDP equation obtained for three values of the parameter $\epsilon$. In top panel, a value $\epsilon = \frac {1} {3}$ is used, 
  with the hyper parameters $\eta = 7  \times 10^{-4}$, $\omega_{data} = 1$, and $\omega_{\mathcal{F}} = 1  \times 10^{-4}$, and 
  $N_c = 48$. In middle panel, a value $\epsilon = 1$ is used, 
  with the hyper parameters $\eta = 7  \times 10^{-4}$, $\omega_{data} = 1$, and $\omega_{\mathcal{F}} = 1  \times 10^{-4}$, and 
  $N_c = 60$. In bottom panel, a value $\epsilon = 5$ is used, 
  with the hyper parameters $\eta = 7  \times 10^{-4}$, $\omega_{data} = 1$, and $\omega_{\mathcal{F}} = 1  \times 10^{-5}$, and 
  $N_c = 70$
   }
\label{fig18}
\end{figure}

\section{Discussion and conclusion}
In this work, we have reviewed the basic concepts of using neural networks in order to integrate differential equations. More specifically, we have focused
on the use of the equations knowledge to penalize the convergence of the training process, and generally referred as physics-informed neural networks
in the literature. A tutorial example on a simple ODE is presented, with the aim to illustrate how adding a partial loss function associated to the physics information
(i.e. via the differential terms ensuring the equation) can considerably ameliorate the results of a normal neural network.

Benchmark tests on different second order ODEs are used in order to highlight the benefits and drawbacks of this approach when compared to a
traditional numerical integration method. When the problem equation displays weak non linearity, the training procedure is successful using known data
representing only the initial conditions (as for a classical integration method). The first advantage of PINNs in this case, is the need to use a very low number
of collocation data points. Indeed, for the problems illustrated in this work, between $20$ and $40$ points are sufficient. An integration using a Runge-Kutta
method (of order two) for the same equations would require a number of points higher by at least one order of magnitude. The second advantage is that, once trained
the solution for a given time (case of an ODE) is instantaneously predicted. This is not the case for a classical integration for which a new integration procedure
must be realized. However, when the non linearity is increased (ses VDP oscillator tests with increasing $\epsilon$ parameter), the knowledge of a higher amount
of training data is required with also a higher number of collocation data points. The previously cited benefits of PINNs are consequently reduced. Nevertheless, the second
benefit remains. The other drawbacks of the method concern the lack of general automatic procedure for a fine tuning of the hyperparameters in order to have an optimal
convergence during the training. Nevertheless, the most interesting point of the PINNs is its meshfree property, contrary to traditional integration methods.

This approach remains relatively recent, and many ameliorations are expected in the next years. In this work, we have focused on the original PINNs method based
on the automatic differentiation to evaluate the equation terms at collocation data points. There is already a wealth of variants to do it differently in order to improve the efficiency of the optimization procedure.
As a very incomplete list of NN-based ideas to solve PDEs, we have methods based on, learning the solution map, and variational formulation (Ritz and Galerkin).
In the context of solving differential equations, PINNs approach can be also used
for other different problems. For example, it is useful for inverse problems when data are known at some time different from the initial conditions that are desired (Raissi et al. 2019). Indeed,
a neural network does not make difference between the two boundaries of the time interval. Second, PINNs technique can be also used when some terms of the differential equations are
not completely known, and which can be added as additional inputs in the input layer.
The aim of the latter problems concerns the data-driven discovery of the governing equations (Rudy et al. 2017).

\begin{acknowledgments}{}
Hubert Baty thanks Emmanuel Franck, Victor Michel-Dansac, and Vincent Vigon (IRMA, Strasbourg), for associating him to the supervision of the Master2 internship
of Vincent Italiano in February-July 2022, which also gave him want to learn the PINNs technique.
\end{acknowledgments}

\end{document}